\documentclass[times, 10pt,twocolumn]{article} 
\usepackage{latex8}
\usepackage{times}
\usepackage{gensymb}
\usepackage{graphicx}
\usepackage{pgfplots}
\usepackage{pgfplotstable}

\usepackage{caption}
\usepackage{hyperref}

\newcommand\blfootnote[1]{%
  \begingroup
  \renewcommand\thefootnote{}\footnote{#1}%
  \addtocounter{footnote}{-1}%
  \endgroup
}

\usepackage{fancyhdr}
\fancypagestyle{firstpage}{
  \fancyhf{}
  \fancyhead[C]{To appear in \textit{Proc. IEEE Multimedia Information Processing and Retrieval (MIPR)}, San Jose, CA, Mar. 2019}
}

\begin{document}

\title{FDDB-360: Face Detection in 360-degree Fisheye Images}

\author{Jianglin Fu, Saeed Ranjbar Alvar, Ivan V. Baji\'{c}, and Rodney G. Vaughan\\
\textit{School of Engineering Science, Simon Fraser University}\\ \textit{Burnaby, BC, V5A 1S6, Canada}\\ \textit{\{jfa49, saeedr, ibajic, rodney\_vaughan\}@sfu.ca}\\
}

\maketitle
\thispagestyle{firstpage}

\begin{abstract}
   360\degree~cameras offer the possibility to cover a large area, for example an entire room, without using multiple distributed vision sensors. However, geometric distortions introduced by their lenses make computer vision problems more challenging. In this paper we address face detection in 360\degree~fisheye images. We show how a face detector trained on regular images can be re-trained for this purpose, and we also provide a 360\degree~fisheye-like version of the popular FDDB face detection dataset, which we call FDDB-360.
\end{abstract}

\textbf{Keywords --} face detection, 360\degree~images, deep learning, FDDB-360 dataset

\Section{Introduction}

Face\blfootnote{This work was supported in part by the NSERC Strategic Project Grant
STPGP 494209} detection is a poster problem of computer vision, with applications in surveillance, security, biometrics, human-computer interaction, and other areas. With the recent advances in deep learning, the problem of face detection in conventional 2D images has largely been solved. For example, on the well-known Face Detection Data Set and Benchmark (FDDB)~\cite{fddbTech}, recent models such as~\cite{DSFD, SFD} reach near 100\% true positive rate with very few false positives. However, face detection in other signal domains, such as 360\degree~images,   point clouds~\cite{face_ptcloud}, or compressed bitstreams~\cite{Alvar_2018_2, Alvar_2018_1}, has been less explored. 

Our focus in this paper is on face detection in 360\degree~images. One way to accomplish face detection in this case would be to project the 360\degree~image onto a set of 2D images, and then employ one of the conventional face detection models on these 2D images. An approach similar to this, where an equirectangular panorama image (a post-processed form of a 360\degree~image) is used for 2D projection and subsequent  object detection, has been presented in~\cite{Yang_2018}. However, the focus of~\cite{Yang_2018} is generic object detection rather than face detection.   Another recent work~\cite{Lee_2018} proposes to generate a spherical image and adjust convolution kernels and pooling operators to work in spherical coordinates, so that Convolutional Neural Network (CNN)-based detectors could operate directly on spherical coordinates.

\begin{figure}[t]
    \centering
    \includegraphics[width=0.42\textwidth]{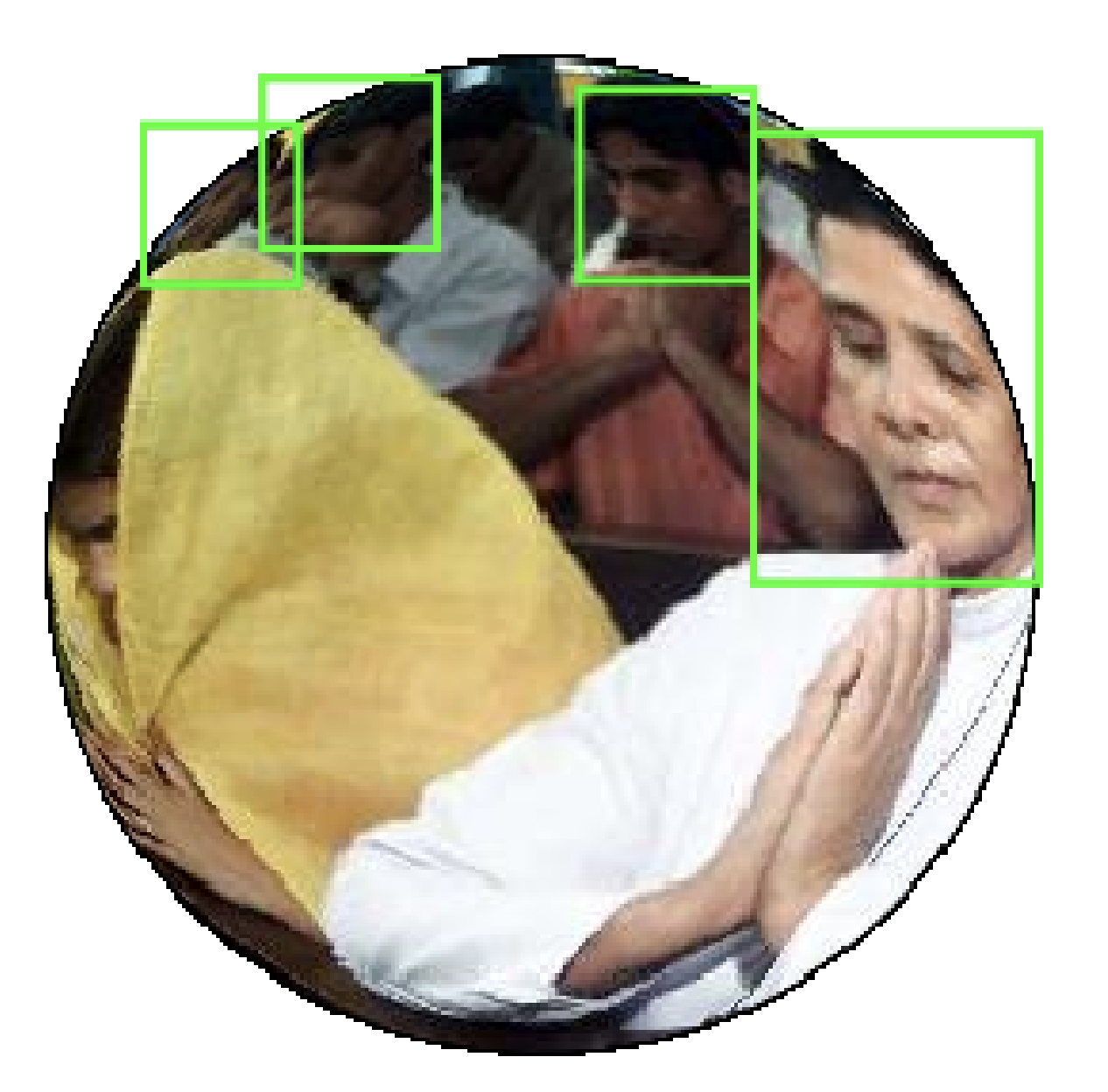}
    \caption{Sample image from our FDDB-360 dataset.}
    \label{fig:sample_image}
\end{figure}

Popular dual-lens 360\degree~cameras (e.g., Ricoh Theta, Samsung Gear 360, Insta 360, etc.) have two hemispherical (``fisheye'') lenses, each projecting a 180\degree~view onto an imaging sensor. The images read directly from the sensors are circular in appearance, and have strong barrel distortion towards the perimeter of the circle. Both the equirectangular panorama used in~\cite{Yang_2018} and the spherical-coordinate image used in~\cite{Lee_2018} are obtained from these fisheye images via post-processing. Clearly, it would be much more efficient to perform face/object detection directly on fisheye images; this would enable the detector to operate closer to the sensor and circumvent unnecessary processing steps. However, training face/object detectors on fisheye images is challenging because of the lack of annotated datasets of fisheye images. We address this challenge by creating a fisheye-looking version of FDDB (which we call FDDB-360) and training a face detector on it. A sample image from FDDB-360 is shown in Fig.~\ref{fig:sample_image}.

The paper is organized as follows. Creation of FDDB-360 from the original FDDB is described in Section~\ref{sec:FDDB-360}. Experiments on FDDB-360 are described in Section~\ref{sec:experiments}, followed by conclusions in Section~\ref{sec:conclusions}. 

\Section{Creating FDDB-360}
\label{sec:FDDB-360}

The purpose of FDDB-360 is to enable the training of models for detecting faces in fisheye images. To this end, we used the annotated images from the original FDDB dataset~\cite{fddbTech} and from them created a number of images that have the appearance of fisheye images. The face locations in the original FDDB are specified by ellipses, but we first converted them to rectangles to enable easier processing.



Fisheye images have least geometric distortions near the center. As we move towards the perimeter, the degree of distortions increases. Face detectors trained on conventional 2D images would likely be able to detect faces near the center of a fisheye image, but detection will become more challenging away from the center. For this reason, we wanted to create fisheye-looking images that would have sufficient number of faces away from the center, so that the detector can learn the appearance of distorted faces. The steps taken to create the new dataset are described in the following subsections. 


\subsection{Image extrapolation}
\begin{figure}[t]
    \centering
    \includegraphics[width=0.49\textwidth]{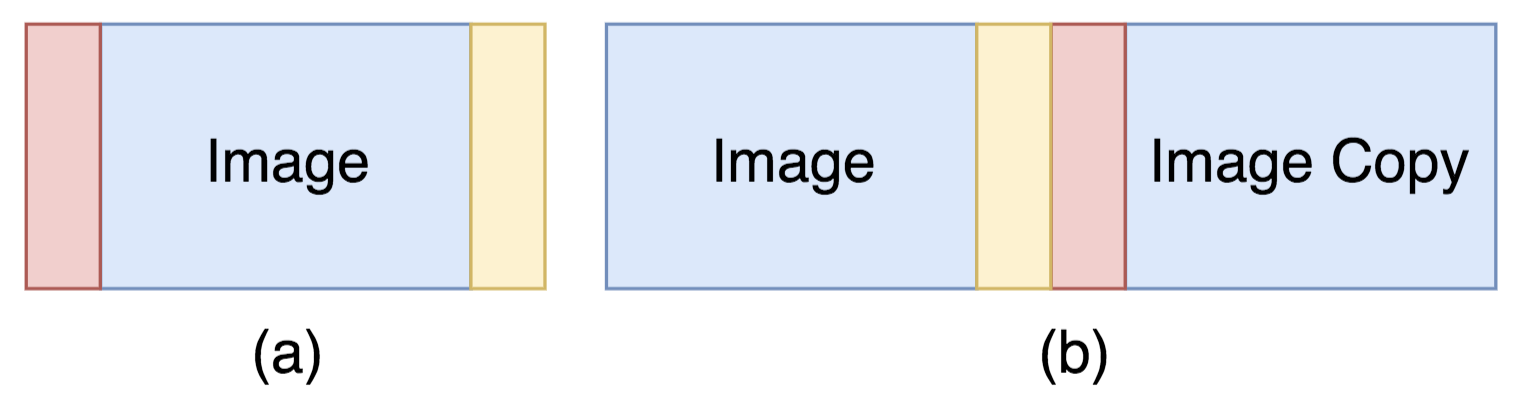}
    \caption{Image extrapolation: (a) Final extrapolated image. (b) Extrapolation carried out as interpolation between two copies of the image. 
    }
    \label{fig:process}
\end{figure}

A number of images in the original FDDB have faces near the center. To facilitate sampling the images with patches where the face locations could be arbitrary, we widened all images by 40\%, by extending it 20\% on both left and right as shown in Fig.~\ref{fig:process}(a). This requires image extrapolation, which is inherently difficult due to the absence of boundary conditions on three sides of the extensions. 

To overcome this challenge, we used the strategy illustrated in Fig.~\ref{fig:process}(b). We created a copy of the image on the right side of the original, spaced away by 40\% of the original width, and then interpolated between the two copies of the image. Though still challenging, this is an easier problem than extrapolation in Fig.~\ref{fig:process}(a) because of a larger area where boundary conditions exist. After interpolation, the interpolated area is split in half, and the right half of it (shown in red in Fig.~\ref{fig:process}) is moved to the left side of the image, to complete the extrapolated image.

Interpolation is carried out using the inpainting algorithm from~\cite{Bajic_2014}, which is an extension of the well-known Criminisi \textit{et al.} method~\cite{Criminisi_2004}. While performing inpainting, we exclude face and skin regions from being used for inpainting. This is because, if the inpainted region ends up with patches containing partial human face, it might confuse the model during training, whether or not these partial faces are actually annotated as faces. To avoid this, while performing inpainting, we increase the cost~\cite{Bajic_2014, Criminisi_2004} of a patch for any patch overlapping with an annotated face, or where skin color is detected~\cite{skin_detection}. 

The approach mentioned above was applied to the majority of images in FDDB. However, about 34\% of FDDB images have width-to-height ratio of less than 3:4. For such images, even 40\% width extension still gives a relatively narrow image. In these cases, we did not rearrange the extended image as shown in Fig.~\ref{fig:process}(a), but left it in the format shown in Fig.~\ref{fig:process}(b), with a copy of the original image on the right side.

\subsection{Fisheye-like distortion}

\begin{figure}[t]
    \centering
    \includegraphics[width=0.35\textwidth]{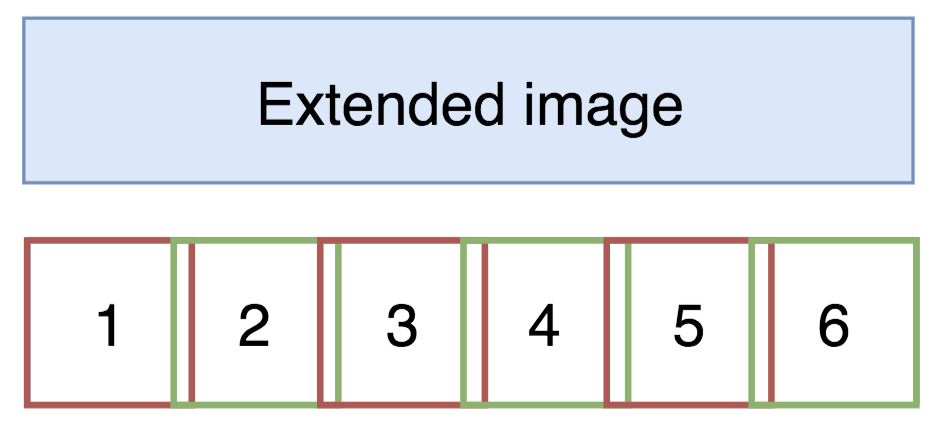}
    \caption{Six square patches, evenly spaced along the width, are extracted from an extended image. 
    }
    \label{fig:crop_img}
\end{figure}

Each extended image is sampled using square patches, which are evenly distributed along its width as shown in  Fig.~\ref{fig:crop_img}. In total, six square patches are extracted from each extended image. Subsequently, fisheye-like distortion is applied to each square patch.

Fisheye distortion models usually involve intrinsic camera parameters\footnote{\url{https://docs.opencv.org/3.4/db/d58/group__calib3d__fisheye.html}} and various lens distortion parameters~\cite{fisheye_distortion_models}. To avoid making distortions camera- and lens-specific, we adopted a simpler approach.

Consider a square patch extracted from an extended image, as shown in Fig.~\ref{fig:crop_img}. We first map this square patch to a circular patch. Let $(x,y)$ be the normalized coordinates of the square patch, such that the patch center is $(0,0)$ and the four corners have coordinates $(\pm1,\pm1)$. The square patch is mapped to a circular patch using the following coordinate mapping:\footnote{\url{https://www.xarg.org/2017/07/how-to-map-a-square-to-a-circle/}} 
\begin{equation}
    (x', \  y') = \left( x\sqrt{1-\frac{y^2}{2}}, \   y\sqrt{1-\frac{x^2}{2}} \right).
    \label{eq:circle_map}
\end{equation}

Such mapping introduces radial distortion, where straight lines get bent towards the perimeter of the circle. To add barrel distortion, which manifests itself as ``squeezing'' towards the perimeter, we further scale the coordinates by a factor that decreases towards the perimeter. Specifically, the mapping is     
\begin{equation}
    (x'',\ y'') = \left(x'e^{-{r^2/4}}, \ y'e^{-{r^2/4}} \right),
    \label{eq:radial_squeezing}
\end{equation}
where $r=\sqrt{(x')^2 + (y')^2}$ is the radial distance from the center of the patch. This form of exponential squeezing was chosen empirically to visually approximate the appearance of fisheye images. An example of a square patch converted to a circular patch with fisheye-like distortion is shown in Fig.~\ref{fig:circular_mapping}.

\subsection{Annotation}

\begin{figure}[t]
    \centering
    \includegraphics[width=0.49\textwidth]{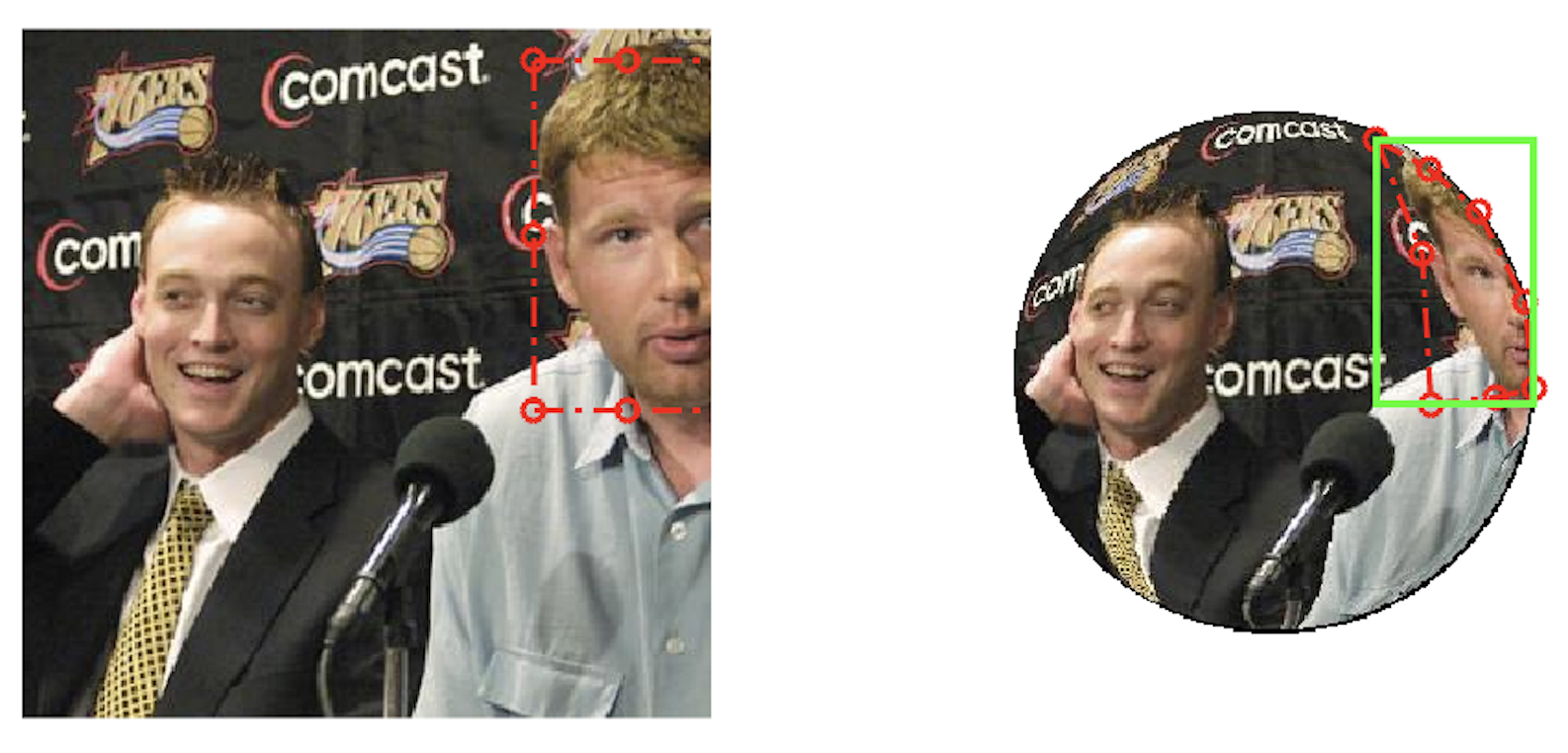}
    \caption{Square patch (\textbf{left}) converted to a circular patch with fisheye-like distortion (\textbf{right}). 
    }
    \label{fig:circular_mapping}
\end{figure}

Once square patches are converted to circular patches, face locations have to be appropriately mapped to the new coordinates. As mentioned earlier, the location of each face in the original FDDB is specified by an ellipse, but we converted those locations to rectangles to simplify further mapping to the circular patch. 

We selected eight points from the bounding box for each face (four corners and four edge midpoints), as illustrated in red in the left part of  Fig.~\ref{fig:circular_mapping}. If a part of the bounding box fell out of the boundaries of the square patch, we trimmed its coordinates to coincide with the patch boundary. Once the mapping (\ref{eq:circle_map})-(\ref{eq:radial_squeezing}) is applied to these eight points, they are mapped to the squeezed circular coordinates as shown in the left part of Fig.~\ref{fig:circular_mapping}. While it is possible to store these eight points (or more, if higher precision is needed) as the bounding polygon for the face in circular coordinates, we decided to simplify the annotations and again use bounding boxes. Hence, we selected the minimum bounding rectangle of the polygon as the annotation for the face location. This is illustrated by a green rectangle in the right part of Fig.~\ref{fig:circular_mapping}. 


When extracting square patches from the extended image, it is possible that faces are cropped and that only a part of the original face falls into the square patch. If the overlap between the original bounding box and the part that is inside the square patch is over 50\%, we kept that annotation and mapped it to the circular patch as explained above. Otherwise, we treated the face as incomplete and did not convert the corresponding annotation to the circular patch.  
\SubSection{Further details of FDDB-360}
After applying the procedures described above, we ended up with  17,052 fisheye-looking images and a total of 26,640 annotated faces. Face locations are provided as the bounding box parameters  \((x, y, w, h)\), where \((x, y)\) is the location of the top-left corner and \((w, h)\) are the width and height of the bounding box, respectively.

Note that the bounding box need not be fully contained within the circular patch, especially for faces that are near the perimeter, as illustrated in the right part of Fig.~\ref{fig:circular_mapping}. From these coordinates, one can find the intersection of the bounding box and the circle if a more precise localization of the face is required. 

The FDDB-360 dataset will be made available online at \url{http://www.sfu.ca/~ibajic/#data}



\begin{figure}[t]
    \centering
    \includegraphics[width=0.42\textwidth]{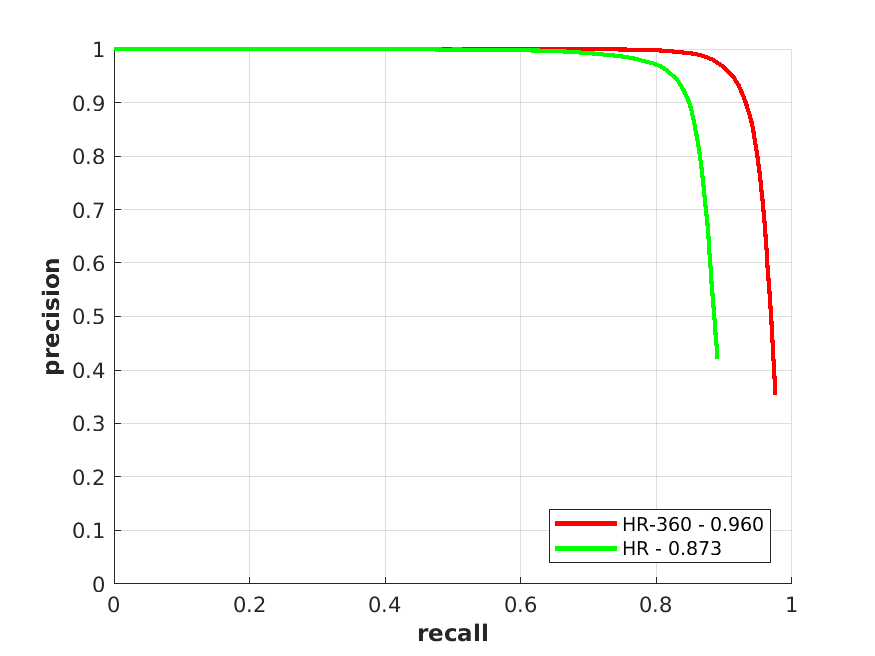}
    \caption{Precision-recall curves on FDDB-360 for HR and HR-360.}
    \label{fig:precision_recall}
\end{figure}
\begin{figure}[t]
    \centering
    \includegraphics[width=0.42\textwidth]{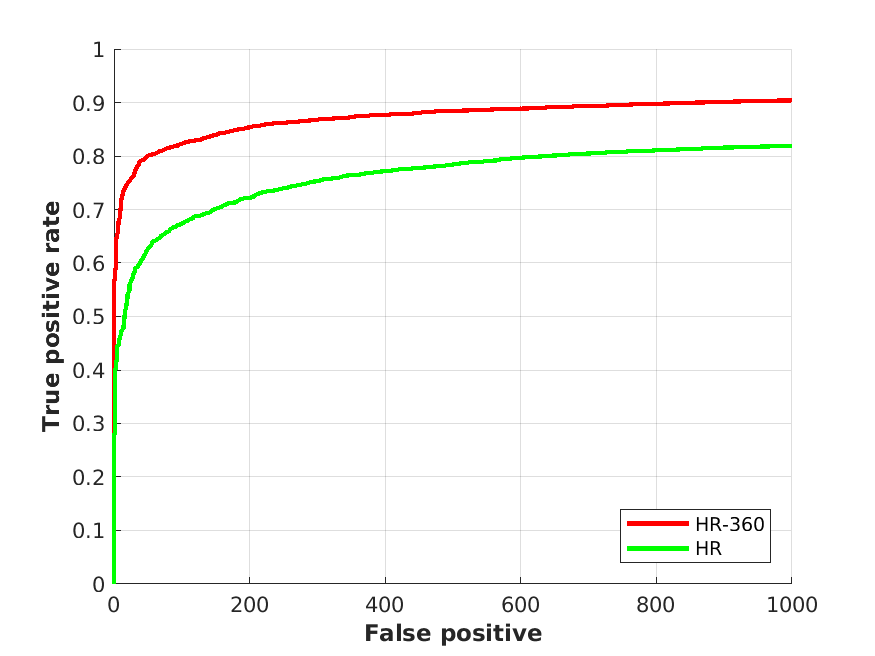}
    \caption{True positive rate versus the number of false positives for HR and HR-360 on FDDB-360.}
    \label{fig:ROC}
\end{figure}

\Section{Experiments}
\label{sec:experiments}
In this section we illustrate the benefits of FDDB-360 by showing that one can re-train an existing face detection model to better detect faces in fisheye-looking images. Our baseline face detector is the well-known TinyFace~\cite{Hu_2017_CVPR}, specifically their hybrid-resolution (HR) model, which was trained on the Wider face dataset~\cite{yang2016wider}. We used their pre-trained model and tested it on FDDB-360 shown. The precision-recall curve of the HR model is shown as the green line in Fig.~\ref{fig:precision_recall}, while its ROC curve is shown as the green line in Fig.~\ref{fig:ROC}. 

Although TinyFace HR is one of the best face detectors on 2D images according to the results on the original FDDB\footnote{\url{http://vis-www.cs.umass.edu/fddb/results.html}}, its performance on FDDB-360 is not particularly impressive. This is understandable, as the model has not been trained on images that involve fisheye-like distortions. Examination of the model's predictions shows that indeed, near the center of the circular patches, the detection performance is quite good, but it degrades towards the perimeter where the  distortions are stronger. 


To show the improvement that can be obtained by transfer learning and re-training, we kept the original ten folds from the FDDB dataset. Experiments were run using 5-fold cross-validation obtained by merging pairs of the original folds as shown in Table~\ref{tab:data_dist}. In each experimental run, we start with the pre-trained model whose weights were obtained on the Wider face dataset~\cite{yang2016wider}. We then further train the model using the software provided by~\cite{Hu_2017_CVPR}, with a learning rate of $10^{-4}$ on one training set shown in Table~\ref{tab:data_dist} and test on the corresponding test/validation set. 

To account for various orientations and scales of faces that may be found in real fisheye images, we performed data augmentation during training. This included horizontal flipping and re-scaling already implemented in the software provided by~\cite{Hu_2017_CVPR}, as well as newly introduced random rotation by 90\degree, 180\degree~and 270\degree. We considered using other rotation angles as well. However, FDDB-360 contains annotations in the form of  axis-aligned rectangles (top left x, top left y, width, height) specifying the location of the face. Rotation by angles other than 90\degree, 180\degree~or 270\degree would cause an increase of the resulting axis-aligned enclosing rectangle, and we felt the ground-truth rectangles were already large enough (see Fig.~\ref{fig:circular_mapping}) and should not be increased further. One possibility to increase the set of viable angles in augmentation in the future would be to switch to a polygonal representation of face locations. 



We refer to the resulting, re-trained model, as HR-360. Red lines in Figs.~\ref{fig:precision_recall}-\ref{fig:ROC} represent the average of five performance curves of HR-360 from the 5-fold cross-validation. 

\begin{table}[]
\centering
\begin{tabular}{|l|l|} \hline
Training & Test/Validation \\ \hline
\hline
Folds 3-10 & Folds 1-2 \\ \hline
Folds 1-2,5-10 & Folds 3-4 \\ \hline
Folds 1-4,7-10 & Folds 5-6 \\ \hline
Folds 1-6,9-10 & Folds 7-8 \\ \hline
Folds 1-8 & Folds 9-10 \\ \hline
\end{tabular}
\caption{\label{tab:data_dist}
Training-test split for 5-fold cross-validation}
\end{table}

As seen in the figures, re-training with transfer learning from the original HR results in significant performance improvement. On the precision-recall results (Fig.~\ref{fig:precision_recall}), HR achieves the area under the curve (AUC) of 0.873, whereas HR-360 achieves 0.960, as indicated in the figure legend. On the true positives (TP) vs. false positives (FP) test (Fig.~\ref{fig:ROC}), HR-360 achieves around 0.85 TP rate with 200 false positives, while the original HR achieves around 0.71 TP rate for the same number of false positives. The difference of about 0.1 TP rate persists for higher number of false positives.

While HR-360 clearly outperforms HR on FDDB-360 overall, we did find some cases where a face was detectable by HR but not by HR-360. Apparently, transfer learning involves some ``forgetting'' as well, and while the model gets new capabilities during transfer learning, some of its old capabilities may disappear. One way around it could be to randomly insert the data that the model was originally trained on (in our case, the Wider face dataset~\cite{yang2016wider}) into the re-training process. However, in keeping with the FDDB-style evaluation, we did not do that, and only used the data from the FDDB-360 folds (Table~\ref{tab:data_dist}) for re-training. 

\begin{figure}[t]
    \centering
    \includegraphics[width=0.5\textwidth]{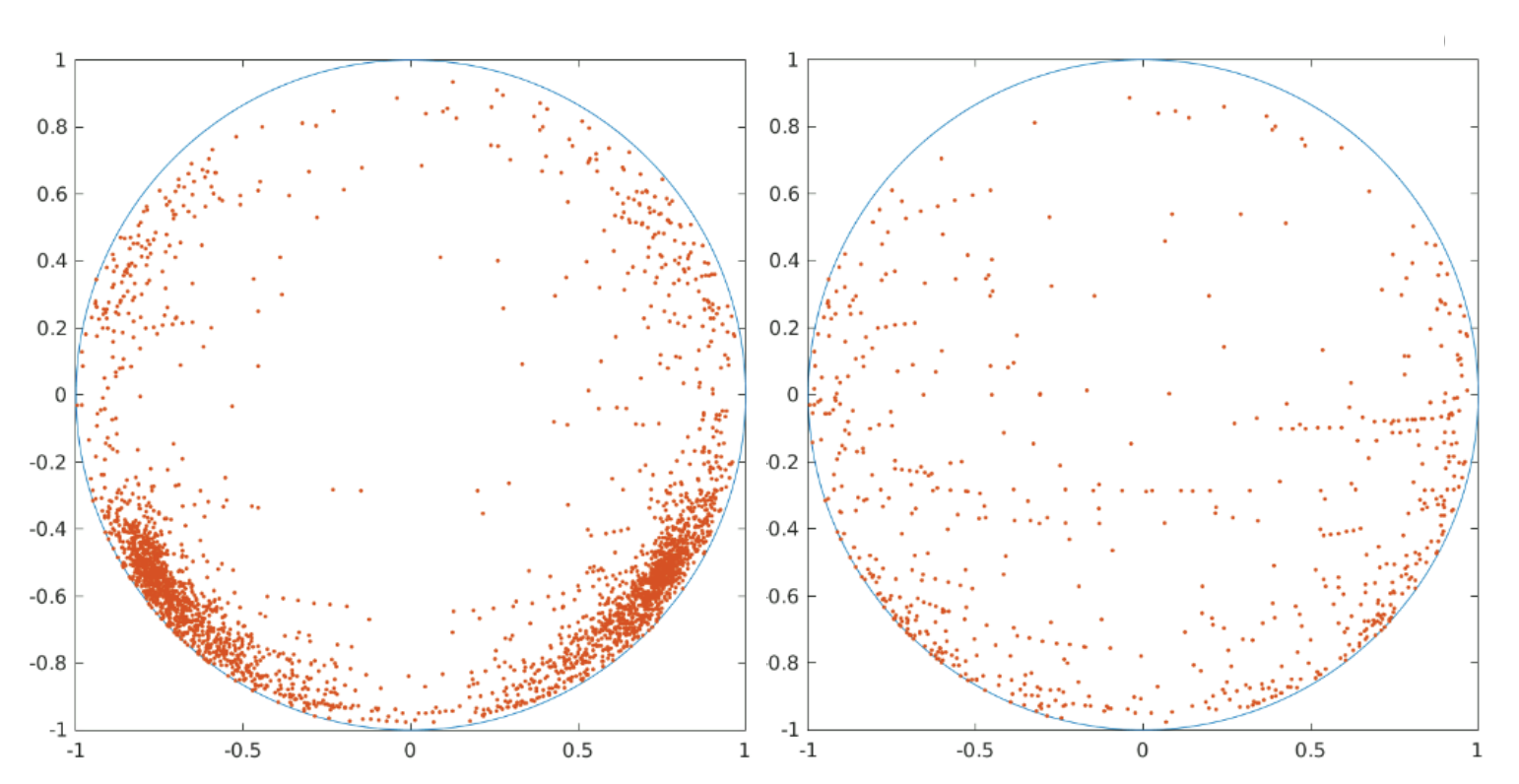}
    \caption{False negative distribution for HR (\textbf{left}) and HR-360 (\textbf{right}).}
    \label{fig:comparison_fn}
    \vspace{-0.5cm}
\end{figure}

To visualize how accurate are HR and HR-360 depending on the location of the target face, we test both models on the entire FDDB-360 dataset and record the locations of all false negatives (FN) - the faces that were missed. This is one way to find how a model can be improved. 
When a face is missed, we record the location of the center of its ground-truth rectangle and normalize it in such a way that the radius of the circular image is 1 (i.e., the image becomes a unit circle). 
In cases where the center of the bounding rectangle is outside the unit circle, the intersection point of the circle and the line connecting the center points of the bounding rectangle and the circle is used to represent the FN location. The scatter plot of FN points for HR and HR-360 are shown in  Fig.~\ref{fig:comparison_fn}. The left graph shows FN's of HR and it is apparent that HR misses many faces near the perimeter of the image, as we expected due to geometric distortions. Meanwhile, the FN's of HR-360 (Fig.~\ref{fig:comparison_fn} right) are more evenly distributed across the unit circle, indicating that the model has learned the corresponding geometric distortions.

\begin{figure}[t]
    \centering
    \includegraphics[width=0.5\textwidth]{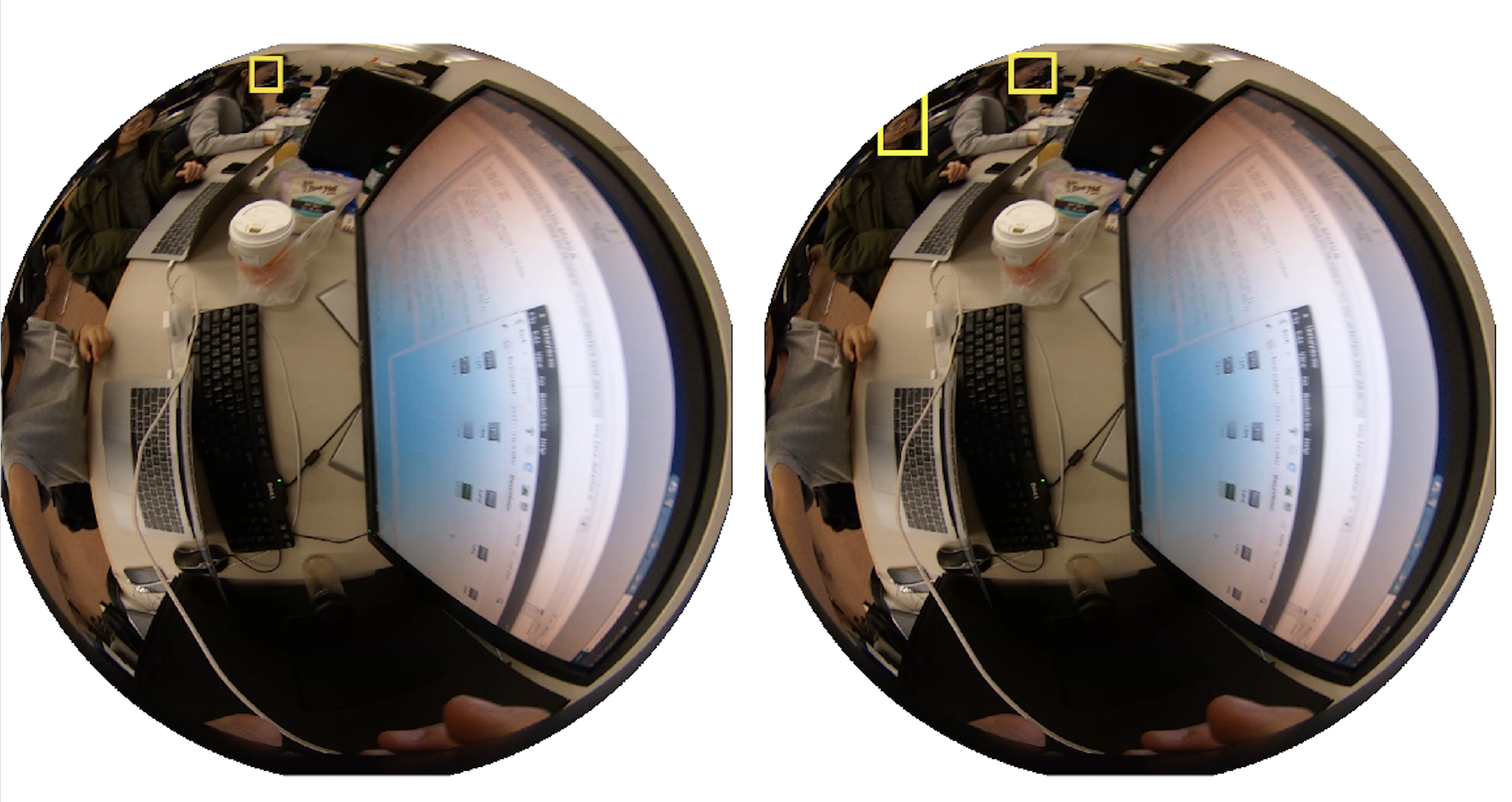}
    \caption{Face detection results by HR (\textbf{left}) and HR-360 (\textbf{right}) on a real fisheye image with several faces along the perimeter. Detected faces are indicated by yellow rectangles.
    }
    \label{fig:demo_HR}
\end{figure}

Finally, we show an example of detection performance on a real fisheye image, rather than an image from FDDB-360. Fig.~\ref{fig:demo_HR} shows an image obtained by a Ricoh Theta V camera with several faces along the perimeter. Image on the left shows the results of HR, where one face was detected, as indicated by the yellow rectangle. Image on the right shows the result of HR-360, which manages to find two faces. There is one more person in the scene (in the left part of the image), whose face is so much out of view that both detectors fail to find it.     

\Section{Conclusions}
\label{sec:conclusions}
In this paper, we described the creation of FDDB-360, a dataset for face detection in fisheye images. The dataset was created from the well-known FDDB dataset by sampling patches from its images and applying fisheye-looking distortion to them, while mapping the face annotations to the new coordinate system. We also showed that re-training (using transfer learning) an existing face detector on FDDB-360 is able to significantly improve its face detection performance on this kind of images. 


\bibliographystyle{latex8}
\bibliography{references}

\end{document}